\documentclass[twoside]{article}
\usepackage[accepted]{aistats2014}
\usepackage{graphicx} 
\usepackage{float,pifont}

\usepackage[belowskip=-6pt,aboveskip=5pt]{caption}

\usepackage{array}
\newcolumntype{L}[1]{>{\raggedright\let\newline\\\arraybackslash\hspace{0pt}}m{#1}}
\newcolumntype{C}[1]{>{\centering\let\newline\\\arraybackslash\hspace{0pt}}m{#1}}
\newcolumntype{R}[1]{>{\raggedleft\let\newline\\\arraybackslash\hspace{0pt}}m{#1}}

\usepackage{color,natbib}
\usepackage{enumerate,multirow}
\usepackage{amsmath,amsthm,amssymb,amsbsy,amsfonts,url}

\newtheorem{thm}{Theorem}
\newtheorem{cor}[thm]{Corollary}
\newtheorem{lem}[thm]{Lemma}
\newtheorem*{nocor}{Corollary}
\newtheorem*{nolem}{Lemma}

\begin{document}

\runningtitle{Student-$t$ Processes as Alternatives to Gaussian Processes}

\runningauthor{Shah, Wilson, Ghahramani }

\twocolumn[

\aistatstitle{Student-$t$ Processes as Alternatives to Gaussian Processes}

\aistatsauthor{ Amar Shah \And Andrew Gordon Wilson \And Zoubin Ghahramani }

\aistatsaddress{ University of Cambridge \And University of Cambridge \And University of Cambridge } ]

\begin{abstract}
We investigate the Student-$t$ process as an alternative to the Gaussian process as a nonparametric prior over functions.  We derive closed form expressions for the marginal likelihood and 
predictive distribution of a Student-$t$ process, by integrating away an inverse Wishart process 
prior over the covariance kernel of a Gaussian process model.  We show surprising 
equivalences between different hierarchical Gaussian process models leading to Student-$t$ processes,
and derive a new sampling scheme for the inverse Wishart process, which helps elucidate these 
equivalences.  Overall, we show that
a Student-$t$ process can retain the attractive properties of a Gaussian process -- a nonparametric
representation, analytic marginal and predictive distributions, and easy model selection through
covariance kernels -- but has enhanced flexibility, and predictive covariances that,
unlike a Gaussian process, explicitly depend on the values of training observations.  We verify
empirically 
that a Student-$t$ process is especially useful in situations where there are changes in 
covariance structure, or in applications like Bayesian optimization, where accurate predictive 
covariances are critical for good performance.  These advantages come at no 
additional computational cost over Gaussian processes. 
\end{abstract}

\section{INTRODUCTION}
Gaussian processes are rich distributions over functions, which provide a Bayesian nonparametric approach to regression. 
Owing to their interpretability, non-parametric flexibility, large support, consistency, simple exact 
learning and inference procedures, and impressive empirical performances \citep{carlthesis}, Gaussian processes as 
kernel machines have steadily grown in popularity over the last decade. 

At the heart of every Gaussian process (GP) is a parametrized covariance kernel, which determines the properties of 
likely functions under a GP.  Typically simple parametric kernels, such as the Gaussian (squared 
exponential) kernel are used, and its parameters are determined through marginal likelihood maximization, having analytically 
integrated away the Gaussian process. However, a fully Bayesian nonparametric treatment of regression would place a nonparametric 
prior over the Gaussian process covariance kernel, to represent uncertainty over the kernel function, and to reflect the natural intuition that the 
kernel does not have a simple parametric form.

Likewise, given the success of Gaussian processes kernel machines, it is also natural to consider more general families of elliptical processes \citep{fangkotzng}, 
such as Student-$t$ processes, where any collection of function values has a desired elliptical distribution, with a covariance matrix constructed 
using a kernel. 

As we will show, the Student-$t$ process can be derived by placing an inverse Wishart process prior on the kernel of a Gaussian process. Given their intuitive value, it is not surprising that various forms of Student-$t$ processes have been used in different applications \citep{robustt,multitaskt,sparset,multiplegps}.  However, the connections between these models, and the theoretical properties of these models, 
remain largely unknown.  Similarly, the practical utility of such models remains uncertain.  For example, \cite{gps} wonder whether  
``the Student-$t$ process is perhaps not as exciting as one might have hoped''.

In short, our paper answers in detail many of the ``what, when and why?'' questions one might have about Student-$t$ processes (TPs), inverse Wishart processes,
and elliptical processes in general.  Specifically: 
\begin{itemize}
 \item We precisely define and motivate the inverse Wishart process \citep{dawid} as a prior over covariance matrices of arbitrary size.
 \item We propose a Student-$t$ process, which we derive from hierarchical Gaussian process models.  We derive 
       analytic forms for the marginal and predictive distributions of this process, and analytic derivatives
       of the marginal likelihood.  
 \item We show that the Student-$t$ process is the most general elliptically symmetric process with analytic marginal 
       and predictive distributions.
 \item We derive a new way of sampling from the inverse Wishart process, which intuitively resolves the seemingly 
       bizarre marginal equivalence between inverse Wishart and inverse Gamma priors for covariance kernels in hierarchical
       GP models.
 \item We show that the predictive covariances of a TP depend on the values of training observations, even though
       the predictive covariances of a GP do not.
 \item We show that, contrary to the Student-$t$ process described in \cite{gps}, an analytic TP noise model can 
       be used which separates signal and noise analytically. 
 \item We demonstrate non-trivial differences in behaviour between the GP and TP on a variety of applications.  We 
       specifically find the TP more robust to change-points and model misspecification, to have notably improved
       predictive covariances, to have useful ``tail-dependence'' between distant function 
       values (which is orthogonal to the choice of kernel), and to be particularly promising for Bayesian optimization, where
       predictive covariances are especially important. 
\end{itemize}

We begin by introducing the inverse Wishart process in section \ref{sec: iwp}. We then derive a 
Student-$t$ process by using an inverse Wishart process over covariance kernels (section \ref{sec: tp}), and 
discuss the properties of this Student$-t$ process in section \ref{sec: proptp}.  Finally, we demonstrate the
Student-$t$ process on regression and Bayesian optimization problems in section \ref{sec: applications}.

\section{INVERSE WISHART PROCESS}
\label{sec: iwp}

In this section we argue that the inverse Wishart distribution is an attractive choice of prior for covariance matrices of arbitrary size. The Wishart distribution is a probability distribution over $\Pi(n)$, the set of real valued, $n \times n$, symmetric, positive definite matrices. Its density function is defined as follows.

{\bf{Definition.}} A random $\Sigma \in \Pi(n)$ is \textit{Wishart} distributed with parameters $\nu >n-1$, $K \in \Pi(n)$, and we write $\Sigma \sim \mathrm{W}_n(\nu, K)$ if its density is given by
\begin{equation}
p(\Sigma) = c_n(\nu, K) |\Sigma|^{(\nu-n-1)/2}
\exp{\Big(-\frac{1}{2}\mathrm{Tr} \big(K^{-1} \Sigma \big)\Big)}, 
\end{equation}
where $c_n(\nu, K) = \Big( |K|^{\nu/2} 2^{\nu n/2} \Gamma_n(\nu/2) \Big)^{-1}$ . \\

The Wishart distribution defined with this parameterization is consistent under marginalization. If $\Sigma \sim \mathrm{W}_n(\nu, K)$, then any $n_1 \times n_1$ principal submatrix $\Sigma_{11}$ is $\mathrm{W}_{n_1}(\nu, K_{11})$ distributed. This property makes the Wishart distribution appear to be an attractive of prior over covariance matrices. Unfortunately the Wishart distribution suffers a flaw which makes it impractical for nonparametric Bayesian modelling.

Suppose we wish to model a covariance matrix using $\nu^{-1}\Sigma$, so that its expected value $\mathbb{E}[\nu^{-1}\Sigma]=K$, and 
$\mathrm{var}[\nu^{-1} \Sigma_{ij}] = \nu^{-1}(K_{ij}^2+K_{ii}K_{jj})$.  Since we require $\nu > n-1$, we must let 
$\nu \rightarrow \infty$ to define a process which has positive semidefinite Wishart distributed marginals of arbitrary size. However, as 
$\nu  \rightarrow \infty $, $\nu^{-1}\Sigma$ tends to the constant matrix $K$ almost surely.  Thus the requirement $\nu>n-1$ prohibits 
defining a useful process which has Wishart marginals of arbitrary size. 
Nevertheless, the \textit{inverse Wishart} distribution does not suffer this problem. \Citet{dawid} parametrized the inverse Wishart distribution as follows:

{\bf{Definition.}} A random $\Sigma \in \Pi(n)$ is \textit{inverse Wishart} distributed with 
parameters $\nu \in \mathbb{R}_{+}$, $K \in \Pi(n)$ and we write $\Sigma \sim \mathrm{IW}_n(\nu, K)$ if its density is given by
\begin{equation}
p(\Sigma) = c_n(\nu, K) |\Sigma|^{-(\nu+2n)/2}
\exp{\Big(-\frac{1}{2}\mathrm{Tr} \big(K \Sigma^{-1} \big)\Big)}, 
\label{iwplik}
\end{equation}
with $c_n(\nu, K) = \dfrac{|K|^{(\nu+n-1)/2}}{2^{(\nu+n-1)n/2} \Gamma_n((\nu+n-1)/2)}$. \\

If $\Sigma \sim \mathrm{IW}_n(\nu, K)$, $\Sigma$ has mean and covariance only when $\nu > 2$ and 
$\mathbb{E}[\Sigma] = (\nu-2)^{-1}K$. Both the Wishart and the inverse Wishart distributions place 
prior mass on every $\Sigma \in \Pi(n)$. Furthermore $\Sigma \sim \mathrm{W}_n(\nu,K)$ if and only 
if $\Sigma^{-1} \sim \mathrm{IW}_n(\nu-n+1,K^{-1})$.   

\Citet{dawid} shows that the inverse Wishart distribution defined as above is consistent under marginalization. 
If $\Sigma \sim \mathrm{IW}_n(\nu, K)$, then any principal submatrix $\Sigma_{11}$ will be $\mathrm{IW}_{n_1}(\nu, K_{11})$ 
distributed. Note the key difference in the parameterizations of both distributions: the parameter $\nu$ does not need to depend 
on the size of the matrix in the inverse Wishart distribution. These properties are desirable and motivate defining a 
process which has inverse Wishart marginals of arbitrary size. Let $\mathcal{X}$ be some input space and 
$k:\mathcal{X} \times \mathcal{X} \rightarrow \mathbb{R}$ a positive definite kernel function. 

{\bf{Definition.}} $\sigma$ is an \textit{inverse Wishart process} on $\mathcal{X}$ with parameters $\nu \in \mathbb{R}_{+}$ and base kernel $k:\mathcal{X} \times \mathcal{X} \rightarrow \mathbb{R}$ if for any finite collection $x_1,...,x_n \in \mathcal{X}$, $\sigma(x_1,...,x_n) \sim \mathrm{IW}_n(\nu, K)$ where $K \in \Pi(n)$ with $K_{ij}=k(x_i,x_j)$. We write $\sigma \sim \mathrm{\mathcal{IWP}}(\nu,k)$.

In the next section we use the inverse Wishart process as a nonparametric prior over kernels in a hierarchical
Gaussian process model.

\section{DERIVING THE STUDENT-$t$ PROCESS}
\label{sec: tp}

\begin{figure}
\begin{minipage}{0.236\textwidth}
 \includegraphics[trim =27mm 213mm 122mm 20mm,clip,width=\textwidth]{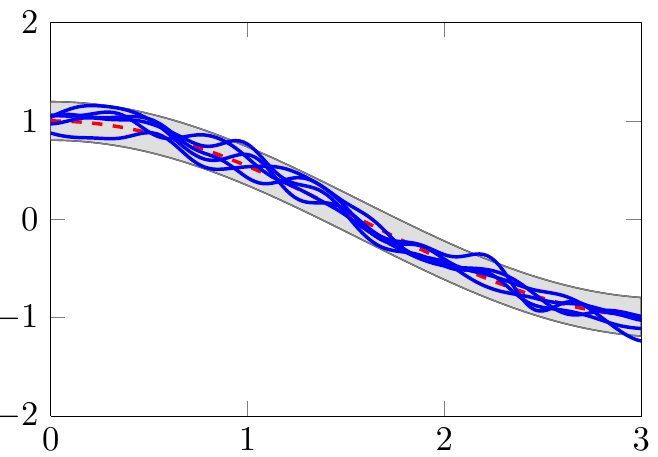}
\end{minipage}
\begin{minipage}{0.236\textwidth}
 \includegraphics[trim =27mm 213mm 122mm 20mm,clip,width=\textwidth]{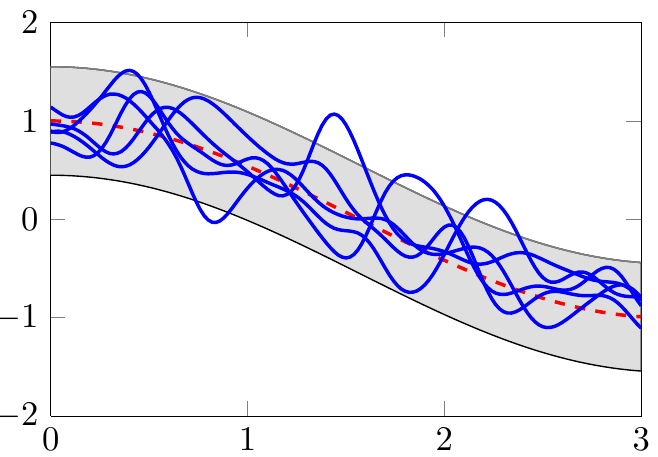}
\end{minipage}
\caption{Five samples (blue solid) from $\mathrm{\mathcal{GP}}(h,\kappa)$ (left) and $\mathrm{\mathcal{TP}}(\nu,h,\kappa)$ (right), with $\nu=5$, $h(x)=\cos (x)$ (red dashed) and $\kappa(x_i,x_j) = 0.01 \exp(-20(x_i-x_j)^2)$. The grey shaded area represents a 95\% predictive interval under each model.  }
\label{priordraws}
\end{figure}

Gaussian processes (GPs) are popular nonparametric Bayesian distributions over functions. 
A thorough guide to GPs has been provided by \citet{gps}. GPs are characterized by a mean 
function and a kernel function. Practitioners tend to use parametric kernel functions and 
learn their hyperparameters using maximum likelihood or sampling based methods. We propose 
placing an inverse Wishart process prior on the kernel function, leading to a Student-$t$
process.

For a base kernel $k_\theta$ parameterized by $\theta$, and a continuous mean function 
$\phi:\mathcal{X} \rightarrow \mathbb{R}$, our generative approach is as follows
\vspace{-3mm}
\begin{align}
\label{gen}
\sigma &\sim \mathrm{\mathcal{\mathcal{IWP}}}(\nu,k_\theta ) \notag \\
y|\sigma &\sim \mathrm{\mathcal{GP}}(\phi, (\nu - 2)\sigma) \,.
\end{align} 

Since the inverse Wishart distribution is a conjugate prior for the covariance matrix 
of a Gaussian likelihood, we can analytically marginalize $\sigma$ in the generative 
model of \eqref{gen}. For any collection of data $\boldsymbol{y} = (y
_1,...,y_n)^\top$ with $\boldsymbol{\phi} = (\phi(x_1),...,\phi(x_n))^\top$,
\begin{align}
p(\boldsymbol{y}|&\nu, K) = \int p( \boldsymbol{y}| \Sigma) p(\Sigma |\nu,K ) d \Sigma \notag \\
&\propto \int \frac{\exp{\bigg(-\frac{1}{2}Tr\Big(\Big(K + \frac{(\boldsymbol{y}-\boldsymbol{\phi})(\boldsymbol{y}-\boldsymbol{\phi})^\top}{\nu-2}\Big) \Sigma^{-1}\Big)\bigg)}}{|\Sigma|^{(\nu+2n+1)/2}} d\Sigma \notag \\
&\propto \Big(1+\frac{1}{\nu-2}(\boldsymbol{y}-\boldsymbol{\phi})^\top K^{-1}(\boldsymbol{y}-\boldsymbol{\phi}) \Big)^{-(\nu+n)/2}
\label{marg_out_IWP}
\end{align}

\vspace{-2mm}
{\bf{Definition.}} $\boldsymbol{y}\in \mathbb{R}^n$ is \textit{multivariate Student-$t$} distributed with parameters $\nu \in \mathbb{R}_{+}\backslash [0,2]$, $\boldsymbol{\phi} \in \mathbb{R}^n$ and $K \in \Pi(n)$ if it has density 
\begin{align}
p(\boldsymbol{y}) &=  \frac{ \Gamma(\frac{\nu+n}{2}) }{ ((\nu-2)\pi)^{ \frac{n}{2}} \Gamma(\frac{\nu}{2})} |K|^{-1/2} \notag \\
&\hspace{10mm} \times \Big(1+\frac{(\boldsymbol{y}-\boldsymbol{\phi})^\top K^{-1}(\boldsymbol{y}-\boldsymbol{\phi})}{\nu-2} \Big)^{-\frac{\nu+n}{2}}
\label{mvtdensity}
\end{align}
We write $\boldsymbol{y} \sim \mathrm{MVT}_n( \nu, \boldsymbol{\phi}, K)$.

We easily compute the mean and covariance of the MVT using the generative derivation:  
$\mathbb{E}[\boldsymbol{y}] = \mathbb{E}[\mathbb{E}[\boldsymbol{y}|\Sigma]]= \boldsymbol{\phi}$ and $\mathrm{cov}[\boldsymbol{y}] = \mathbb{E}[\mathbb{E}[(\boldsymbol{y}-\boldsymbol{\phi})(\boldsymbol{y}-\boldsymbol{\phi})^\top|\Sigma]] = \mathbb{E}[(\nu-2)\Sigma] =K$. We prove the following Lemma in the Supplementary Material.

\begin{lem}
The multivariate Student-$t$ is consistent under marginalization.
\label{mvtmarg}
\end{lem}

We define a Student-$t$ process as follows. 

{\bf{Definition.}} $f$ is a \textit{Student-$t$ process} on $\mathcal{X}$ with parameters $\nu >2$, mean function $\Psi:\mathcal{X}\rightarrow \mathbb{R}$, and kernel function $k:\mathcal{X} \times \mathcal{X} \rightarrow \mathbb{R}$ if any finite collection of function values have a joint multivariate Student-$t$ distribution, i.e. $(f(x_1),...,f(x_n))^\top \sim \mathrm{MVT}_n(\nu, \boldsymbol{\phi}, K)$ where $K \in \Pi(n)$ with $K_{ij}=k(x_i,x_j)$ and $\phi \in \mathbb{R}^n$ with $\phi_i = \Phi(x_i)$. We write $f \sim \mathrm{\mathcal{TP}}(\nu,\Phi,k)$.

\section{TP PROPERTIES \& RELATION TO OTHER PROCESSES}
\label{sec: proptp}

In this section we discuss the conditional distribution of the TP, the relationship between GPs and TPs, 
another covariance prior which leads to the same TP, elliptical processes, and a sampling scheme for the IWP 
which gives insight into this equivalence.  Finally we consider modelling noisy functions with a TP.

\subsection{Relation to Gaussian process}

The Student-$t$ process generalizes the Gaussian process. A GP can be seen as a limiting case of a TP as shown in Lemma \ref{nuinfinity}, which is proven in the Supplementary Material. 

\begin{lem}
Suppose $f \sim \mathrm{\mathcal{TP}}(\nu,\Phi,k)$ and $g \sim \mathrm{\mathcal{GP}}(\Phi,k)$. Then $f$ tends to $g$ in distribution as $\nu \rightarrow \infty$.
\label{nuinfinity}
\end{lem}

The $\nu$ parameter controls how \textit{heavy tailed} the process is. Smaller values of $\nu$ correspond to heavier tails. As $\nu$ gets larger, the tails converge to Gaussian tails. This is illustrated in prior sample draws shown in Figure \ref{priordraws}. Notice that the samples from the TP tend to have more extreme behaviour than the GP.

\begin{figure}
\centering
\includegraphics[trim =27mm 195mm 20mm 25mm, clip, width = 0.485\textwidth, page = 3]{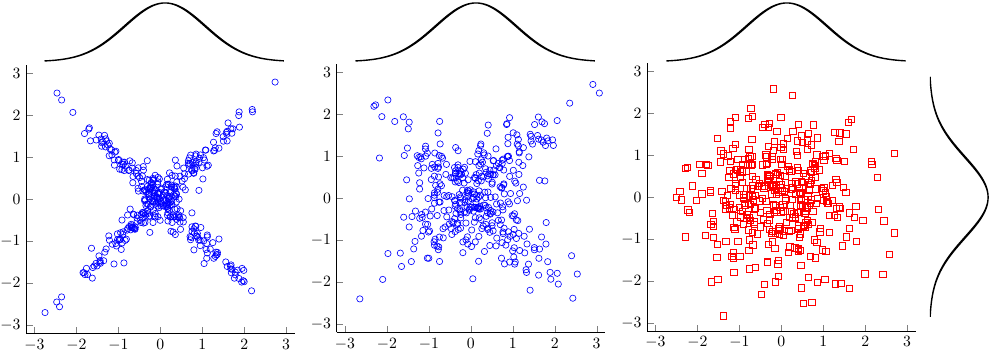}
\caption{Uncorrelated bivariate samples from a Student-$t$ copula with $\nu=3$ (left), a Student-$t$ copula with $\nu=10$ (centre) and a Gaussian copula (right). All marginal distributions are $\mathrm{N}(0,1)$ distributed.}
\label{copulas}
\end{figure}

$\nu$ also controls the nature of the dependence between variables which are jointly 
Student-$t$ distributed, and not just their marginal distributions. In Figure \ref{copulas} we show plots of 
samples which all have Gaussian marginals but different joint distributions. Notice how the tail dependency of these 
distributions is controlled by $\nu$.  For example, the dependencies between $y(x_p)$ and $y(x_q)$ are different 
depending on whether $y$ is a TP or a GP, even if the TP and GP have the same kernel.

\subsection{Conditional distribution}

The conditional distribution for a multivariate Student-$t$ has an analytic form which we state in Lemma \ref{cond} and prove in the Supplementary Material.

\begin{lem} Suppose $\boldsymbol{y} \sim \mathrm{MVT}_n( \nu, \boldsymbol{\phi}, K)$ and let $\boldsymbol{y_1}$ and $\boldsymbol{y_2}$ represent the first $n_1$ and remaining $n_2$ entries of $\boldsymbol{y}$ respectively. Then
\begin{equation} 
\boldsymbol{y_2}|\boldsymbol{y_1} \sim \mathrm{MVT}_{n_2}\Big(\nu + n_1, \tilde{\boldsymbol{\phi_2}},  \frac{\nu + \beta_1 - 2}{\nu + n_1 -2} \times \tilde{K}_{22} \Big), \label{mvtcond}
\end{equation}
where $\tilde{\boldsymbol{\phi_2}}=K_{21}K_{11}^{-1}(\boldsymbol{y_1}-\boldsymbol{\phi_1})+\boldsymbol{\phi_2}$, $\beta_1 = (\boldsymbol{y_1}-\boldsymbol{\phi_1})^\top K_{11}^{-1}(\boldsymbol{y_1}-\boldsymbol{\phi_1})$ and $\tilde{K}_{22} = K_{22}-K_{21}K_{11}^{-1}K_{12}$. Note that $\mathbb{E}[\boldsymbol{y_2}|\boldsymbol{y_1}] = \tilde{\boldsymbol{\phi_2}}$, 
$\mathrm{cov}[\boldsymbol{y_2}|\boldsymbol{y_1}] = \frac{\nu + \beta_1 - 2}{\nu + n_1 -2} \times \tilde{K}_{22}$.
\label{cond}
\end{lem}

As $\nu$ tends to infinity, this predictive distribution tends to a Gaussian process predictive distribution as we would expect given Lemma \ref{nuinfinity}. Perhaps less intuitively, this predictive distribution also tends to a Gaussian process predictive as $n_1$ tends to infinity. 

The predictive mean has the same form as for a Gaussian process, conditioned on having the same kernel $k$, with the same hyperparameters. The key difference is in the predictive covariance, which now explicitly depends on 
the training observations. Indeed, a somewhat disappointing feature of the Gaussian process is that for a given kernel, the predictive covariance of 
new samples does not depend on training observations. Importantly, since the marginal likelihood of the TP in \eqref{mvtdensity} differs from the marginal likelihood of the GP, both the predictive mean and predictive covariance of a TP will differ from that of a GP, after learning kernel hyperparameters.

The scaling constant of the multivariate Student-$t$ predictive covariance has an intuitive explanation. Note that $\beta_1$ is distributed as the sum of squares of $n_1$ independent $\mathrm{MVT}_1(\nu,0,1)$ distributions and hence $\mathbb{E}[\beta_1] = n_1$. If the observed value of $\beta_1$ is larger than $n_1$, the predictive covariance is scaled up and vice versa. The magnitude of scaling is controlled by $\nu$.

\subsection{Another Covariance Prior}

Despite the apparent flexibility of the inverse Wishart distribution, we illustrate in Lemma \ref{simpleintegral} the surprising result that a multivariate Student-$t$ distribution can be derived using a much simpler covariance prior which has been considered previously \citep{robustt}. The proof can be found in the Supplementary Material.

\begin{lem}
Let $K \in \Pi(n)$, $\boldsymbol{\phi}\in \mathbb{R}^n$, $\nu > 2$, $\rho > 0$ and
\begin{align}
r^{-1} &\sim \Gamma(\nu/2,\rho/2) \notag \\
\boldsymbol{y} | r &\sim \mathrm{N}_n(\boldsymbol{\phi}, r(\nu-2)K/\rho ),
\label{simplesetup}
\end{align}  
then marginally $\boldsymbol{y} \sim \mathrm{MVT}_{n}(\nu,\boldsymbol{\phi},K )$.
\label{simpleintegral}
\end{lem}
From \eqref{simplesetup}, $r^{-1}|\boldsymbol{y} \sim \Gamma \big( \frac{\nu+n}{2}, \frac{\rho}{2}(1+\frac{\beta}{\nu-2})  \big)$ and hence $\mathbb{E}[(\nu-2)r/\rho|\boldsymbol{y}] = \frac{\nu+\beta-2}{\nu+n-2}$. This is exactly the factor by which $\tilde{K}_{22}$ is scaled in the MVT conditional distribution in \eqref{mvtcond}.

This result is surprising because we previously integrated over an infinite dimensional nonparametric object (the IWP) to derive the Student-$t$ process, yet here we show that we can integrate over a single scale parameter (inverse Gamma) to arrive at the same marginal process. We provide some insight into why these distinct priors lead to the same marginal multivariate Student-$t$ distribution in section \ref{newsamp}.

\subsection{Elliptical Processes}

We now show that both Gaussian and Student-$t$ processes are \textit{elliptically symmetric}, and that the Student-$t$ process
is the more general elliptical process.

{\bf{Definition.}} $\boldsymbol{y} \in \mathbb{R}^n$ is \textit{elliptically symmetric} if and only if there exists $\boldsymbol{\mu} \in \mathbb{R}^n$, $R$ a nonnegative random variable, $\Omega$ a $n\times d$ matrix with maximal rank $d$ and $\boldsymbol{u}$ uniformly distributed on the unit sphere in $\mathbb{R}^d$ independent of $R$ such that $\boldsymbol{y} \stackrel{\mathcal{D}}{=} \boldsymbol{\mu} + R\Omega \boldsymbol{u}$, where $\stackrel{\mathcal{D}}{=}$ denotes equality in distribution.

An overview of elliptically symmetric distributions and the following Lemma can be found in \citet{fangkotzng}. 

\begin{lem}
Suppose $R_1 \sim \chi^2(n)$ and $R_2 \sim \Gamma^{-1}(\nu/2,1/2)$ independently. If $R = \sqrt{R_1}$, then $\boldsymbol{y}$ is Gaussian distributed. If $R=\sqrt{(\nu-2)R_1 R_2}$  then $\boldsymbol{y}$ is MVT distributed.
\label{ellipse_examples}
\end{lem}

Elliptically symmetric distributions characterize a large class of distributions which are unimodal and where the likelihood of a point decreases in its distance from this mode. These properties are natural assumptions we often want to encode in our prior distribution, making elliptical distributions ideal for multivariate modelling tasks. The idea naturally extends to infinite dimensional objects.

{\bf{Definition.}} Let $\mathcal{Y}=\{y_i\}$ be a countable family of random variables. It is an \textit{elliptical process} if any finite subset of them are jointly elliptically symmetric.

Not all elliptical distributions have densities (e.g.\ L\'{e}vy, alpha-stable distributions). Even fewer elliptical processes have densities, and the set of those that do is characterized in Theorem \ref{kelkerthm} due to \citet{kelker}.

\begin{thm}
Suppose $\mathcal{Y}=\{y_i\}$ is an elliptical process. Any finite collection $\boldsymbol{z}=\{z_1,...,z_n \} \subset \mathcal{Y}$ has a density if and only if there exists a non-negative random variable $r$ such that $\boldsymbol{z}|r \sim \mathrm{N}_n(\boldsymbol{\mu},r\Omega\Omega^\top)$. 
\label{kelkerthm}
\end{thm}

A simple corollary of this theorem describes the only two cases where an elliptical process has an analytically representable density function (its proof is included in the Supplementary Material).

\begin{cor}
Suppose $\mathcal{Y}=\{y_i\}$ is an elliptical process. Any finite collection $\boldsymbol{z}=\{z_1,...,z_n \} \subset \mathcal{Y}$ has an analytically representable density if and only if $\mathcal{Y}$ is either a Gaussian process or a Student-$t$ process.
\end{cor}

Since the Student-$t$ process generalizes the Gaussian process, it is the most general elliptical process which has an analytically representable density. The TP is thus an expressive tool for nonparametric Bayesian modelling. 

With analytic expressions for the predictive distributions, the same computational costs as a Gaussian process and increased flexibility, the Student-$t$ process can be used as a drop-in replacement for a Gaussian process in many applications. 

\subsection{A New Way to Sample the IWP}
\label{newsamp}

We show that the density of an inverse Wishart distribution depends only on the eigenvalues of a positive definite matrix. To the best of our knowledge this change of variables has not been computed previously. This decomposition offers a novel way of sampling from an inverse Wishart distribution and insight into why the Student-$t$ process can be derived using an inverse Gamma or an inverse Wishart process covariance prior. 

Let $\Xi(n)$ be the set of all $n \times n$ orthogonal matrices. A matrix is orthogonal if it is square, real valued and its rows and columns are orthogonal unit vectors. Orthogonal matrices are compositions of rotations and reflections, which are volume preserving operations. Symmetric positive definite (SPD) matrices can be represented through a diagonal and an orthogonal matrix:
\begin{thm}
Let $\Sigma \in \Pi(n)$, the set of SPD, $n \times n$ matrices. Suppose $\{ \lambda_1,...,\lambda_n \}$ are the eigenvalues of $\Sigma$. There exists $Q \in \Xi(n)$ such that $\Sigma = Q \Lambda Q^\top$, where $\Lambda = \mathrm{diag}(\lambda_1,...,\lambda_n )$. 
\label{orthog}
\end{thm}

Now suppose $\Sigma \sim \mathrm{IW}_n(\nu,I)$. We compute the density of an IW using the representation in Theorem \ref{orthog}, being careful to include the Jacobian of the change of variable, $J(\Sigma ;Q,\Lambda)$, given in \cite{RMT}. From \eqref{iwplik} and using the facts that $Q^\top Q = I$ and $|AB|=|BA|$,
\begin{align}
p&(\Sigma) d \Sigma = p(Q\Lambda Q^\top) 
|J(\Sigma ;Q,\Lambda)|
d\Lambda dQ 
\notag \\
&\propto |Q \Lambda Q^\top|^{-(\nu+2n)/2} \exp\Big(-\frac{1}{2}\mathrm{Tr} \big((Q \Lambda Q^\top)^{-1}  \big)\Big) \notag \\
& \hspace{10mm} \times \Big|  Q^\top \prod_{1 \leq i<j \leq n} |\lambda_i - \lambda_j |\Big| 
d\Lambda dQ 
\notag \\
&\propto \prod_{i=1}^n \bigg( \lambda_i^{-\frac{\nu+2n}{2}} e^{-\frac{1}{2\lambda_i}} 
\prod_{j \neq i} |\lambda_i - \lambda_j|^n d \lambda_i \bigg) \hspace{1mm} dQ
\label{evalues}
\end{align}

\eqref{evalues} tells us that $Q$ is uniformly distributed over $\Xi(n)$ (e.g. from a $\Upsilon_{n,n}$ distribution as described in \citet{dawid2}) and that the $\lambda_i$ are exchangeable, i.e., permuting the $\mathrm{diag}(\Lambda)$ does not affect its probability. We denote this exchangeable distribution $\Theta_n(\nu)$. We generate a draw from an inverse Wishart distribution by sampling $Q \sim \Upsilon_{n,n}$, $\Lambda \sim \Theta_n(\nu)$ and setting $\Sigma = Q \Lambda Q^\top$.

This result provides a geometric interpretation of what a sample from $\mathrm{IW}_n(\nu,I)$ looks like. We first uniformly at random pick an orthogonal set of basis vectors in $\mathbb{R}^n$ and then stretch these basis vectors using an exchangeable set of scalar random variables. An analogous interpretation holds for the Wishart distribution. 


Recall from Lemma \ref{ellipse_examples} that if $\boldsymbol{u}$ is uniformly distributed on the unit sphere in $\mathbb{R}^n$ and $R \sim \chi^2(n)$ independently, then $\sqrt{R}\boldsymbol{u} \sim \mathrm{N}_n(0,I)$. By \eqref{marg_out_IWP} and Lemma \ref{ellipse_examples}, if we sample $Q$ and $\Lambda$ from the generative process above, then $\sqrt{(\nu-2)R} Q \Lambda^{1/2} \boldsymbol{u}$ is marginally a draw from $\mathrm{MVT}(\nu,0,I)$. Since the diagonal elements of $\Lambda$ are exchangeable, $Q$ is orthogonal and sampled uniformly over $\Xi(n)$, and $\boldsymbol{u}$ is spherically symmetric, we must have that $Q \Lambda^{1/2} \boldsymbol{u} \stackrel{\mathcal{D}}{=} \sqrt{R'} \boldsymbol{u}$ for some positive scalar random variable $R'$ by symmetry. By Lemma \ref{ellipse_examples} we know $R' \sim \Gamma^{-1}(\nu/2,1/2)$. In summary, the action of $Q \Lambda^{1/2}$ on $\boldsymbol{u}$ is equivalent in distribution to a rescaling by an inverse Gamma variate.

\subsection{Modelling Noisy Functions}

It is common practice to assume that outputs are the sum of a latent Gaussian process and independent Gaussian noise. An advantage of such a model is in the fact that the sum of independent Gaussian distributions is Gaussian distributed and hence such a Gaussian process model remains analytic in the presence of noise. Unfortunately the sum of two independent MVTs is analytically intractable. 

This problem was encountered by \citet{gps}, who went on to dismiss the multivariate Student-$t$ process for practical purposes. Our approach is to incorporate the noise into the kernel function, for example, letting $k = k_{\theta} + \delta$, where $k_{\theta}$ is a parametrized kernel and $\delta$ is a diagonal kernel function. Such a model is not equivalent to adding independent noise, since the scaling parameter $\nu$ will have an effect on the squared-exponential kernel as well as the noise kernel. \cite{multitaskt} propose a similar method for handling noise; however, they incorrectly assume that the latent function and noise are independent under this model. The noise will be uncorrelated with the latent function, but not independent. 

\begin{figure}[t]
\centering
\begin{minipage}{0.16\textwidth}
 \includegraphics[trim =25mm 199mm 130mm 20mm,clip,width=\textwidth]{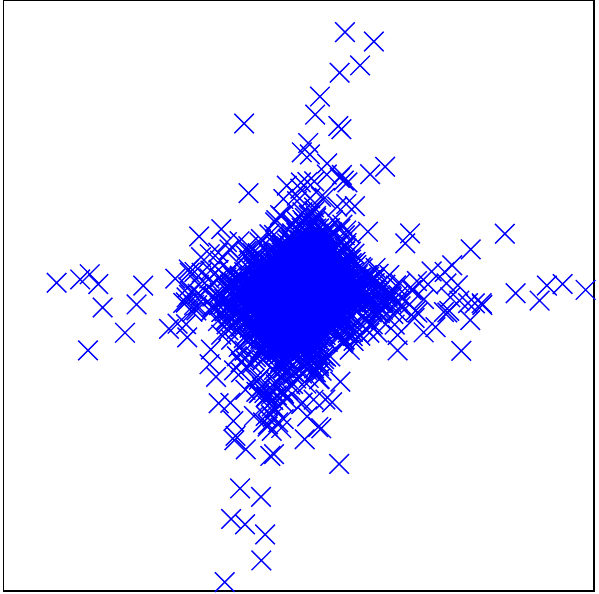}
\end{minipage}
\begin{minipage}{0.16\textwidth}
 \includegraphics[trim =25mm 199mm 130mm 20mm,clip,width=\textwidth]{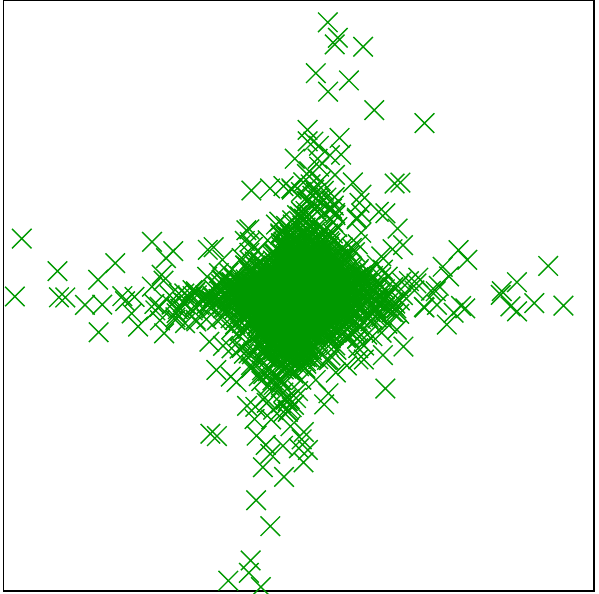}
\end{minipage} 
\begin{minipage}{0.16\textwidth}
 \includegraphics[trim =25mm 199mm 130mm 20mm,clip,width=\textwidth]{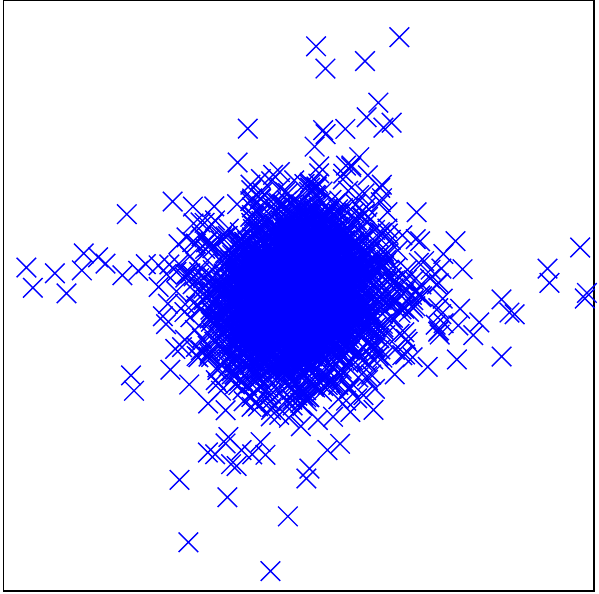}
\end{minipage} 
\begin{minipage}{0.16\textwidth}
 \includegraphics[trim =25mm 199mm 130mm 20mm,clip,width=\textwidth]{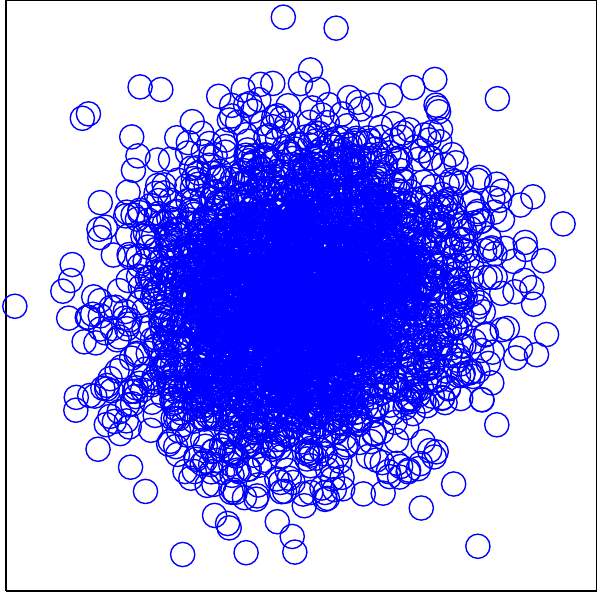}
\end{minipage}
\caption{Scatter plots of points drawn from various 2-dim processes. Here $\nu=2.1$ and $K_{ij} = 0.8\delta_{ij} + 0.2$.
\textbf{Top-left}: $\mathrm{MVT}_2(\nu,0,K) + \mathrm{MVT}_2(\nu,0,0.5I)$.
\textbf{Top-right}: $\mathrm{MVT}_2(\nu,0,K+0.5I)$ (our model).
\textbf{Bottom-left}: $\mathrm{MVT}_2(\nu,0,K) + \mathrm{N}_2(0,0.5I)$.
\textbf{Bottom-right}: $\mathrm{N}_2(0,K+0.5I)$. } \label{samples}
\end{figure}

As $\nu \rightarrow \infty$ this model tends to a GP with independent Gaussian noise. In Figure \ref{samples}, we consider samples from various two dimensional processes when $\nu$ is small and the signal to noise ratio is small. Here we see that the TP with noise incorporated into its kernel behaves similarly to a TP with independent Student-$t$ noise.

There have been several attempts to make GP regression robust to heavy tailed noise that rely on approximate inference \citep{nealtlik,gptlik}. It is hence attractive that our proposed method can model heavy tailed noise whilst retaining an analytic inference scheme. This is a novel finding to the best of our knowledge.

\section{APPLICATIONS}
\label{sec: applications}

\begin{figure}
\begin{minipage}{0.236\textwidth}
 \includegraphics[trim =27mm 193mm 122mm 20mm,clip,width=\textwidth]{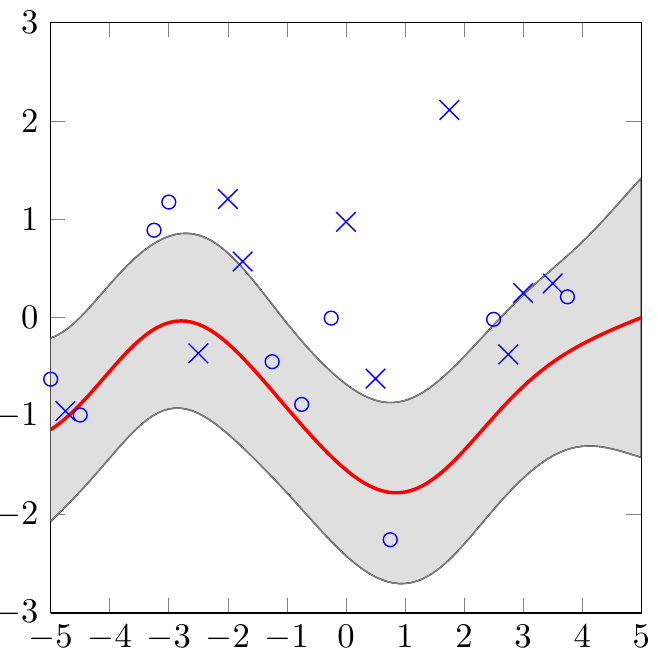}
\end{minipage}
\begin{minipage}{0.236\textwidth}
 \includegraphics[trim =27mm 193mm 122mm 20mm,clip,width=\textwidth]{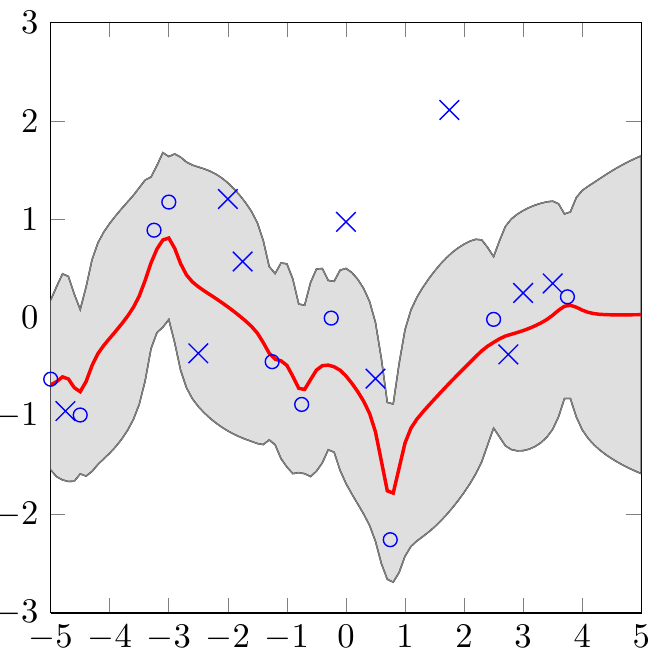}
\end{minipage}
\caption{Posterior distributions of 1 sample from Synthetic Data B under GP prior (left) and TP prior (right). The solid line is the posterior mean, the shaded area represents a 95\% predictive interval,  circles are training points and crosses are test points.}
\label{synthetic}
\end{figure}
In this section we compare TPs to GPs for regression and Bayesian optimization.

\subsection{Regression}

Consider a set of observations $\{x_i,y_i\}_{i=1}^n$ for $x_i \in \mathcal{X}$ and $y_i \in \mathbb{R}$. Analogous to Gaussian process regression, we assume the following generative model
\begin{align}
f &\sim \mathrm{\mathcal{TP}}(\nu, \Phi, k_\theta ) \notag \\
y_i &= f(x_i)   \hspace{5mm} \text{for } i=1,...,n.
\end{align} 

In this work we consider parametric kernel functions. A key task when using such kernels is in learning the parameters of the chosen kernel, which are called the hyperparameters of the model. We include derivatives of the marginal log likelihood of the TP with respect to the hyperparameters in the Supplementary Material.

\subsubsection{Experiments}

We test the Student-$t$ process as a regression model on a number of datasets. We sample hyperparameters using Hamiltonian Monte Carlo \citep{hmc} and use a kernel function which is a sum of a squared exponential and a delta kernel function ($k_\theta = k_\mathrm{SE}$). The results for all of these experiments are summarized in Table \ref{regresults}. 

\begin{table}[h]
\caption{Predictive Mean Squared Errors (MSE) and Log Likelihoods (LL) of regression experiments. The TP consistently has the lowest MSE and highest LL.}
\label{regresults}
\begin{center}
\begin{scriptsize}
\begin{sc}
\setlength{\tabcolsep}{4pt}
\begin{tabular}{lccccc}
\hline
\multirow{2}{*}{Data set}  & \multicolumn{2}{c}{Gaussian Process} & \multicolumn{2}{c}{Student-T Process}  \\
 &  MSE &  LL &  MSE &  LL  \\
\hline
Synth A   &  2.24 $\pm$ 0.09 & -1.66$\pm$ 0.04  & 2.29 $\pm$ 0.08 & \bf{-1.00$\pm$ 0.03}   \\
Synth B   &  9.53 $\pm$ 0.03 & -1.45$\pm$ 0.02  & \bf{5.69 $\pm$ 0.03} & \bf{-1.30$\pm$ 0.02}   \\
Snow     &  10.2 $\pm$ 0.08  &   4.00 $\pm$ 0.12   &  10.5 $\pm$ 0.07   &   \bf{25.7  $\pm$  0.18} \\
Spatial        & 6.89 $\pm$0.04  &  4.34$\pm$0.22   & \bf{5.71 $\pm $0.03} & \bf{44.4$\pm$0.4}   \\
Wine         & 4.84 $\pm$ 0.08 & -1.4 $\pm$ 1 & \bf{4.20 $\pm$ 0.06}  &  \bf{113 $\pm$ 2}  \\
\hline
\end{tabular}
\end{sc}
\end{scriptsize}
\end{center}
\vskip -0.1in
\end{table}

\textbf{Synthetic Data A.} We sample 100 functions from a GP prior with Gaussian noise and fit both GPs and TPs to the data with the goal of predicting test points. For each function we train on 80 data points and test on 20. The TP, which generalizes the GP, has superior predictive uncertainty in this example. 

\textbf{Synthetic Data B.} We construct data by drawing 100 functions from a GP with a squared exponential kernel and adding Student-$t$ noise independently. The posterior distribution of one sample is shown in Figure \ref{synthetic}. The predictive means are also not identical since the posterior distributions of the hyperparameters differ between the TP and the GP. Here the TP has a superior predictive mean, since after hyperparameter training it is better able to model Student-$t$ noise, as well as better predictive uncertainty.

\textbf{Whistler Snowfall Data\footnote{The snowfall dataset can be found at \url{http://www.climate.weatheroffice.ec.gc.ca.}}.} Daily snowfall amounts in Whistler have been recorded for the years 2010 and 2011. This data exhibits clear changepoint type behaviour due to seasonality which the TP handles much better than the GP.  

\textbf{Spatial Interpolation Data\footnote{The spatial interpolation data can be found at \url{http://www.ai_geostats.org} under SIC97.}.} This dataset contains rainfall measurements at 467 (100 observed and 367 to be estimated) locations in Switzerland on 8 May 1986. 

\textbf{Wine Data.} This dataset due to \citet{wine} consists of 12 attributes of various red wines including acidity, density, pH and alcohol level. Each wine is given a corresponding quality score between 0 and 10. We choose a random subset of 400 wines: 360 for training and 40 for testing.

\begin{figure}
\centering
 \includegraphics[trim =26mm 190mm 122mm 20mm,clip,width=0.34\textwidth]{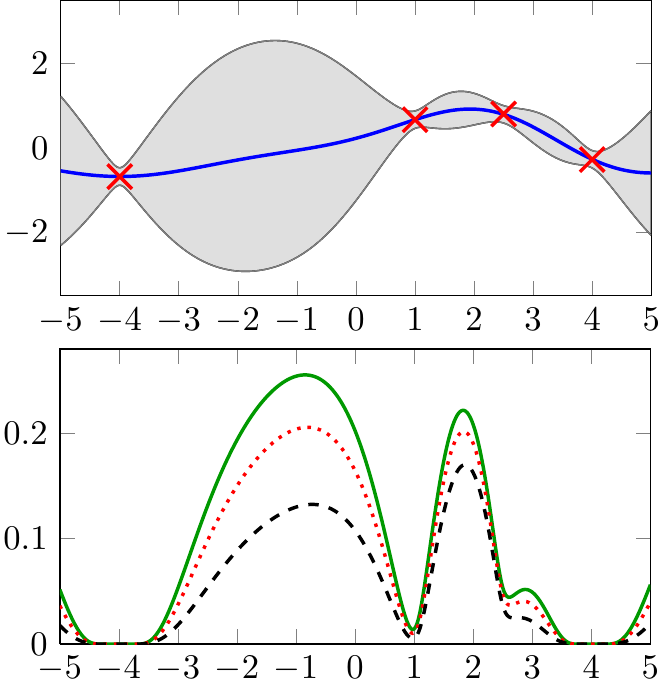}
\caption{Posterior distribution of a function to maximize under a GP prior (top) and acquisition functions (bottom). The solid green line is the acquisition function for a GP, the dotted red and dashed black lines are for TP priors with $\nu = 15$ and $\nu=5$ respectively. All other hyperparameters are kept the same. }
\label{acqui}
\end{figure}

\subsection{Bayesian Optimization}

Machine learning algorithms often require tuning parameters, which control learning rates and abilities, via optimizing an objective function. One can model this objective function using a Gaussian process, under a powerful iterative optimization procedure known as Gaussian process Bayesian optimization \citep{gpboreview2010}. To pick where to query the objective function next, one can optimize the expected improvement (EI) over the running optimum, the probability of improving the current best or a GP upper confidence bound.

\subsubsection{Method}

In this paper we work with the EI criterion and for reasons described in \citet{snoek} we use an ARD Mat\'{e}rn 5/2 kernel defined as
\begin{align}
k_{M52}(\boldsymbol{x},\boldsymbol{x}') &= \theta_0 \Big( 1+ \sqrt{5 r^2_{\boldsymbol{x},\boldsymbol{x}'} } \Big) \exp \Big( - \sqrt{5 r^2_{\boldsymbol{x},\boldsymbol{x}'} } \Big)  
\end{align}
where $r^2(\boldsymbol{x},\boldsymbol{x}') = \sum_{d=1}^D \frac{(x_d-x_d')^2}{\theta_d^2}$.

We assume that the function we wish to optimize over is $f: \mathbb{R}^D \rightarrow \mathbb{R}$ and is drawn from a multivariate Student-$t$ process with scale parameter $\nu>2$, constant mean $\mu$ and kernel function a linear sum of a ARD Mat\'{e}rn 5/2 kernel and a delta function kernel. 
 
Our goal is to find where $f$ attains its minimum. Let $X_N=\{\boldsymbol{x}_n,f_n\}_{n=1}^N$ be our current set of $N$ observations and $f_{\mathrm{best}} = \min \{f_1,...,f_N\}$. To compress notation we let $\boldsymbol{\theta}$ represent the parameters $\theta,\nu,\mu$. Let the acquisition function $a_{\mathrm{EI}}\big(\boldsymbol{\boldsymbol{x}} ; X_N, \boldsymbol{\theta}\big)$ denote the expected improvement over the current best value from choosing to sample at point $\boldsymbol{x}$ given current observations $X_N$ and hyperparameters $\boldsymbol{\theta}$. Note that the distribution of $f(\boldsymbol{x})|X_N,\boldsymbol{\theta}$ is $\mathrm{MVT}_1(\nu+N, \tilde{\mu}(\boldsymbol{x}; X_n), \tilde{\tau}(\boldsymbol{x}; X_n, \nu)^2)$, where the form of $\tilde{\mu}$ and $\tilde{\tau}$ are derived in \eqref{mvtcond}. Let $\tilde{\gamma}=\frac{f_{\mathrm{best}} -\tilde{\mu}}{\tilde{\tau}}$. Then

\begin{figure*}[t]
\centering
\begin{minipage}{0.31\textwidth}
\centering
 \includegraphics[trim =32mm 188mm 95mm 21mm,clip,width=\textwidth]{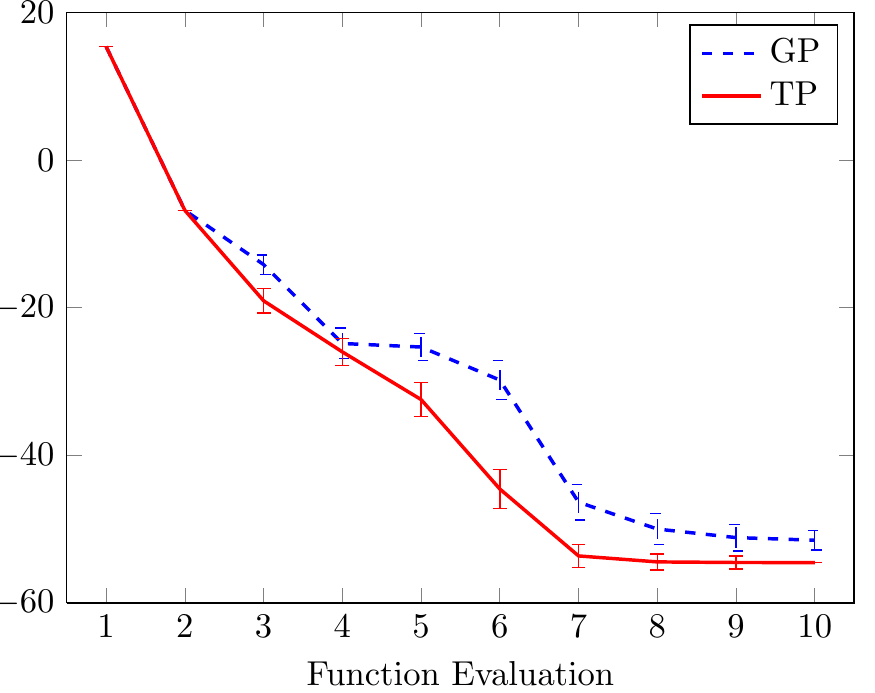}
\end{minipage}
\begin{minipage}{0.31\textwidth}
\centering
 \includegraphics[trim =31mm 188mm 95mm 21mm,clip,width=\textwidth]{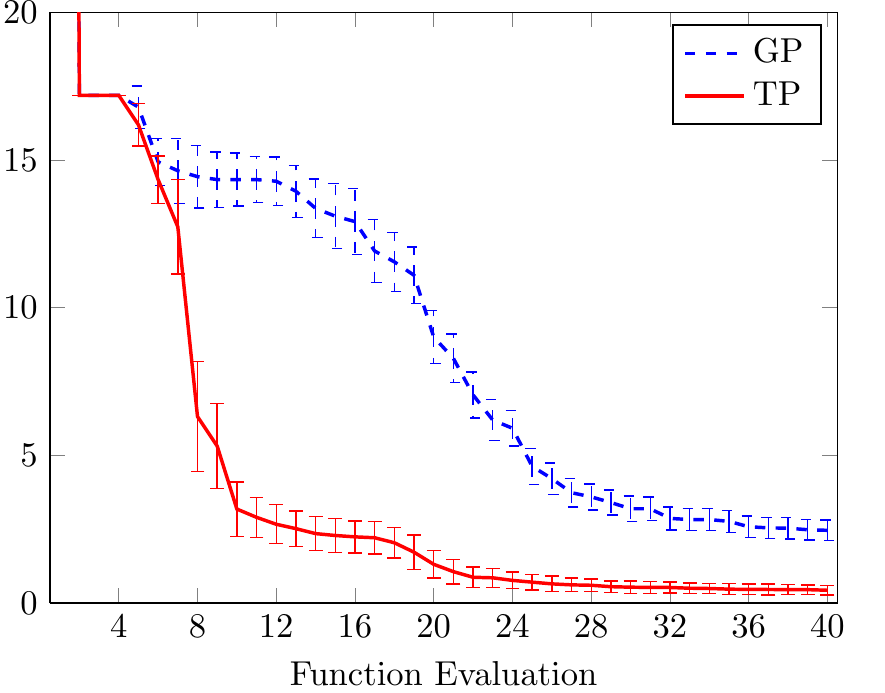}
\end{minipage}
\begin{minipage}{0.31\textwidth}
\centering
 \includegraphics[trim =32mm 188mm 95mm 21mm,clip,width=\textwidth]{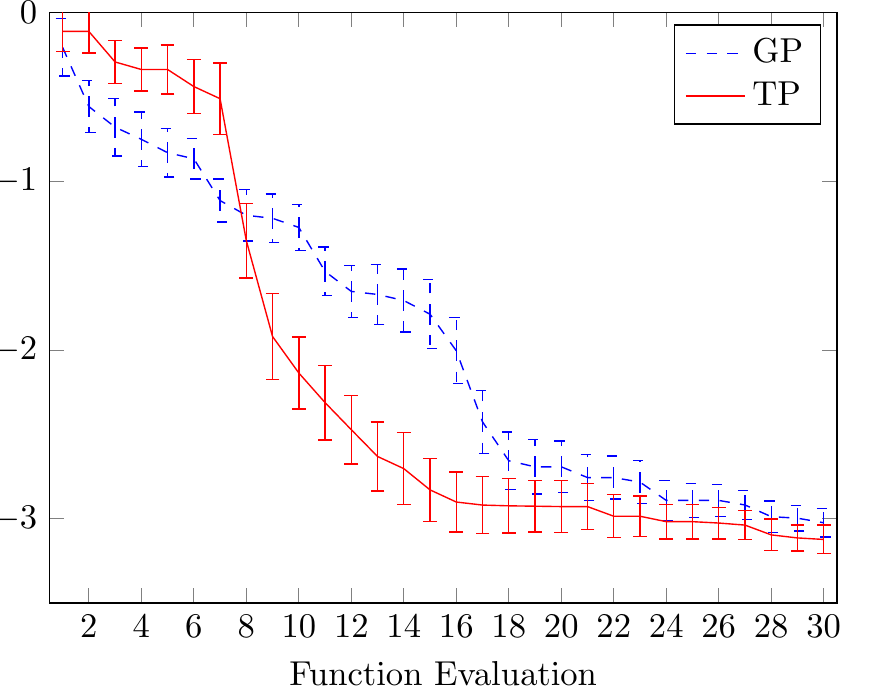}
\end{minipage}
\caption{Function evaluations for the synthetic function (left), Branin-Hoo function (centre) and the Hartmann function (right). Evaluations under a Student-$t$ process prior (solid line) and a Gaussian process prior (dashed line) are shown. Error bars represent the standard deviation of 50 runs. In each panel we are minimizing an objective function. The vertical axis represents the running minimum function value.}
\label{expts}
\end{figure*}

\begin{align}
&a_{\mathrm{EI}}\big(\boldsymbol{\boldsymbol{x}} ; X_N, \boldsymbol{\theta} \big) = \mathbb{E}\big[ \max\big(f_{\mathrm{best}} - f(\boldsymbol{\boldsymbol{x}}),0\big) | X_N, \boldsymbol{\theta} \big]  \notag \\
& \hspace{3mm}= \int_{-\infty}^{f_{\mathrm{best}}}dy (f_{\mathrm{best}} - y) \frac{1}{\tilde{\tau}} \lambda_{\nu + N}\Big( \frac{y- \tilde{\mu}}{\tilde{\tau}} \Big) \notag \\
&\hspace{3mm}= \tilde{\gamma} \tilde{\tau}  \Lambda_{\nu+N}(\tilde{\gamma})
+ \tilde{\tau} \Big(1+ \frac{\tilde{\gamma}^2-1}{\nu+N-1}  \Big)  \lambda_{\nu+N}(\tilde{\gamma}),
\end{align}
where $\lambda_\nu$ and $\Lambda_\nu$ are the density and distribution functions of a $\mathrm{MVT}_1(\nu,0,1)$ distribution respectively.

The parameters $\boldsymbol{\theta}$ are all sampled from the posterior using slice sampling, similar to the method used in \citet{snoek}. Suppose we have $H$ sets of posterior samples $\{ \boldsymbol{\theta}_h \}_{h=1}^H$. We set
\begin{equation}
\tilde{a}_{\mathrm{EI}}\big(\boldsymbol{x} ; X_N \big) = \frac{1}{H}\sum_{h=1}^H a_{\mathrm{EI}}\big(\boldsymbol{x} ; X_N, \boldsymbol{\theta}_h \big)
\end{equation}
as our approximate marginalized acquisition function. The choice of the net place to sample is $\boldsymbol{x}_{\mathrm{next}} = \mathrm{argmax}_{\boldsymbol{x} \in \mathbb{R}^D} \tilde{a}_{\mathrm{EI}}\big(\boldsymbol{x} ; X_N \big) $, which we find by using gradient descent based methods starting from a dense set of points in the input space. 

To get more intuition on how $\nu$ changes the behaviour of the acquisition function, we study an example in Figure \ref{acqui}. Here we fix all hyperparameters other than $\nu$ and plot the acquisition functions varying $\nu$. In this example, it is clear that in certain scenarios the TP prior and GP prior will lead to very different proposals given the same information.

\subsubsection{Experiments}
\vspace{-1mm}

We compare a TP prior with a Mat\'{e}rn plus a delta function kernel to a GP prior with the same kernel, for Bayesian optimization. To integrate away uncertainty we slice sample the hyperparameters \citep{slice}. We consider 3 functions: a 1-dim synthetic sinusoidal, the 2-dim Branin-Hoo function and a 6-dim Hartmann function. All the results are shown in Figure \ref{expts}.

\textbf{Sinusoidal synthetic function} \hspace{3mm} In this experiment we aimed to find the minimum of $f(x) = -(x-1)^2 \sin (3x + 5x^{-1} +1) $ in the interval $[5,10]$. The function has 2 local minima in this interval. TP optimization clearly outperforms GP optimization in this problem; the TP was able to come to within 0.1\% of the minimum in $8.1 \pm 0.4$ iterations whilst the GP took $10.7 \pm 0.6$ iterations.

\textbf{Branin-Hoo function} \hspace{3mm} This function is a popular benchmark for optimization methods \citep{branin} and is defined on the set $\{(x_1,x_2): 0 \leq x_1 \leq 15, -5 \leq x_2 \leq 15 \}$. We initialized the runs with 4 initial observations, one for each corner of the input square.  

\textbf{Hartmann function} \hspace{3mm} This is a function with 6 local minima in $[0,1]^6$ \citep{hartmann}. The runs are initialised with 6 observations at corners of the unit cube in $\mathbb{R}^6$. Notice that the TP tends to behave more like a step function whereas the Gaussian process' rate of improvement is somewhat more constant. The reason for this behaviour is that the TP tends to more thoroughly explore any modes which it has found, before moving away from these modes. This phenomenon seems more prevalant in higher dimensions. 


\section{CONCLUSIONS}
\vspace{-1mm}

We have shown that the inverse Wishart process (IWP) is an appropriate prior over covariance matrices of arbitrary size. We used an IWP prior over a GP kernel and showed that marginalizing over the IWP results in a Student-$t$ process (TP). The TP has consistent marginals, closed form conditionals and contains the Gaussian process as a special case. We also proved that the TP is the only elliptical process other than the GP which has an analytically representable density function.  The TP prior was applied in regression and Bayesian optimization tasks, showing improved performance over GPs with no additional computational costs.

The take home message for practitioners should be that the TP has many if not all of the benefits of GPs, 
but with increased modelling flexibility at no extra cost. Our work suggests that
it could be useful to replace GPs with TPs in almost any application. The added flexibility of the TP is orthogonal to the choice of kernel, and could complement recent expressive closed form kernels \citep{spectralmixkernel,gpatt} in future work.

\bibliographystyle{plainnat}
\bibliography{references}

\appendix
\makeatletter

\makeatother

\onecolumn

\section*{\center{\LARGE{Supplementary Material}}}
\hspace{4mm}

In Appendix \ref{proofs}, we provide proofs of Lemmas and Corollaries from our paper. We describe the derivatives of the log marginal likelihood of the Student-$t$ process which is useful for hyperparameter learning in Appendix \ref{derivs}. In Appendix \ref{equiv} we offer more insights as to why two seemingly different covariance priors for a Gaussian process prior lead to the same marginal distribution. 

\section{Proofs}
\label{proofs}

\vspace{2mm}
\begin{nolem} \textbf{[1]}
The multivariate Student-$t$ is consistent under marginalization.
\begin{proof} 
Assume the generative process of equation 3 of the main text. $\Sigma_{11}$ is $\mathrm{IW}_{n_1}(\nu, K_{11})$ distributed for any principal submatrix of $\Sigma$. Futhermore $y_1|\Sigma_{11} \sim \mathrm{N}_{n_1}(0, (\nu - 2)\Sigma_{11})$ since the Gaussian distribution is consistent under marginalization. Hence $y_1 \sim \mathrm{MVT}_{n_1}( \nu, \mu_1, K_{11})$.
\end{proof}
\end{nolem}

\vspace{2mm}
\begin{nolem} \textbf{[2]}
Suppose $f \sim \mathrm{\mathcal{TP}}(\nu,\Phi,k)$ and $g \sim \mathrm{\mathcal{GP}}(\Phi,k)$. Then $f$ tends to $g$ in distribution as $\nu \rightarrow \infty$.
\begin{proof}
It is sufficient to show convergence in density for any finite collection of inputs. Let $\boldsymbol{y} \sim \mathrm{MVT}_n( \nu, \boldsymbol{\phi}, K)$ and set $\beta = (\boldsymbol{y}-\boldsymbol{\phi})^\top K^{-1}(\boldsymbol{y}-\boldsymbol{\phi})$ then
\begin{align*}
p(\boldsymbol{y}) &\propto \Big(1+\frac{\beta}{\nu-2} \Big)^{-(\nu+n)/2}  \rightarrow e^{-\beta/2}
\end{align*} 
an $\nu \rightarrow \infty$. Hence the distribution of $\boldsymbol{y}$ tends to a $\mathrm{N}_n(\boldsymbol{\phi}, K)$ distribution as $\nu \rightarrow \infty$.
\end{proof}
\end{nolem}

\vspace{2mm}
\begin{nolem}\textbf{[3]}
Suppose $\boldsymbol{y} \sim \mathrm{MVT}_n( \nu, \boldsymbol{\phi}, K)$ and let $\boldsymbol{y_1}$ and $\boldsymbol{y_2}$ represent the first $n_1$ and remaining $n_2$ entries of $\boldsymbol{y}$ respectively. Then
\begin{equation} 
\boldsymbol{y_2}|\boldsymbol{y_1} \sim \mathrm{MVT}_{n_2}\Big(\nu + n_1, \tilde{\boldsymbol{\phi_2}},  \frac{\nu + \beta_1 - 2}{\nu + n_1 -2} \times \tilde{K}_{22} \Big), \label{mvtcond}
\end{equation}
where $\tilde{\boldsymbol{\phi_2}}=K_{21}K_{11}^{-1}(\boldsymbol{y_1}-\boldsymbol{\phi_1})-\boldsymbol{\phi_2}$, $\beta_1 = (\boldsymbol{y_1}-\boldsymbol{\phi_1})^\top K_{11}^{-1}(\boldsymbol{y_1}-\boldsymbol{\phi_1})$ and $\tilde{K}_{22} = K_{22}-K_{21}K_{11}^{-1}K_{12}$.
\begin{proof}
Let $\beta_2 = (\boldsymbol{y_2}-\tilde{\boldsymbol{\phi_2}})^\top \tilde{K}_{22}^{-1}(\boldsymbol{y_2}-\tilde{\boldsymbol{\phi_2}})$. Note that $\beta_1 + \beta_2 = (\boldsymbol{y}-\boldsymbol{\phi})^\top K^{-1}(\boldsymbol{y}-\boldsymbol{\phi})$. We have
\begin{align*}
p(\boldsymbol{y_2}|\boldsymbol{y_1}) = \frac{p(\boldsymbol{y_1},\boldsymbol{y_2})}{p(\boldsymbol{y_1})} &\propto  
\Big(1+\frac{\beta_1+\beta_2}{\nu-2} \Big)^{-(\nu+n)/2}    
\Big(1+\frac{\beta_1}{\nu-2} \Big)^{(\nu+n_1)/2}    
\notag \\
&\propto 
\Big(1+\frac{\beta_2}{\beta_1+\nu-2} \Big)^{-(\nu+n)/2}
\end{align*}
Comparing this expression to the definition of a MVT density function gives the required result.
\end{proof}
\end{nolem}

\vspace{2mm}
\begin{nolem}\textbf{[4]}
Let $K \in \Pi(n)$, $\boldsymbol{\phi}\in \mathbb{R}^n$, $\nu > 2$, $\rho > 0$ and
\begin{align}
r^{-1} &\sim \Gamma(\nu/2,\rho/2) \notag \\
\boldsymbol{y} | r &\sim \mathrm{N}_n(\boldsymbol{\phi}, r(\nu-2)K/\rho ),
\end{align}  
then marginally $\boldsymbol{y} \sim \mathrm{MVT}_{n}(\nu,\boldsymbol{\phi},K )$.
\begin{proof} 
Let $\beta = (\boldsymbol{y}-\boldsymbol{\phi})^\top K^{-1}(\boldsymbol{y}-\boldsymbol{\phi})$. We can analytically marginalize out the scalar $r$,
\begin{align*}
p(\boldsymbol{y}) = \int p( \boldsymbol{y}| r) p(r ) d r 
&\propto \int \exp \Big( - \frac{\rho \beta}{2(\nu-2)r} \Big) r^{-\frac{n}{2}}
\exp \Big( - \frac{\rho}{2r} \Big) r^{-\frac{(\nu+2)}{2}} dr  \\
&\propto \Big( 1 + \frac{\beta}{\nu-2} \Big)^{-\frac{(\nu+n)}{2}}
\int \exp\Big( - \frac{1}{2r} \Big) r^{-\frac{(\nu+n+2)}{2}} dr  \\
&\propto \Big( 1 + \frac{\beta}{\nu-2} \Big)^{-\frac{(\nu+n)}{2}}
\end{align*}
Hence $\boldsymbol{y} \sim \mathrm{MVT}_n(\nu, \boldsymbol{\phi}, K)$ . Note the redundancy in $\rho$. Without loss of generality, let $\rho = 1$.
\end{proof}
\end{nolem}

\vspace{2mm}
\begin{nocor}\textbf{[7]}
Suppose $\mathcal{Y}=\{y_i\}$ is an elliptical process. Any finite collection $\boldsymbol{z}=\{z_1,...,z_n \} \subset \mathcal{Y}$ has an analytically representable density if and only if $\mathcal{Y}$ is either a Gaussian process or a Student-$t$ process.
\begin{proof}
By Theorem 6, we need to be able to analytically solve $\int p(\boldsymbol{z}|r)p(r) dr$, where $\boldsymbol{z}|r \sim \mathrm{N}_n(\boldsymbol{\mu},r\Omega\Omega^\top)$. This is possible either when $r$ is a constant with probability 1 or when $r \sim \Gamma^{-1}(\nu/2,1/2)$, the conjugate prior. These lead to the Gaussian and Student-$t$ processes respectively.
\end{proof}
\end{nocor}

\section{Marginal Likelihood Derivatives}
\label{derivs}

Being able to analytically compute the derivative of the likelihood with respect to the hyperparameters is useful for hyperparameter learning e.g. maximum likelihood or Hamiltonian (Hybrid) Monte Carlo.
\begin{align*}
\log &p(\boldsymbol{y} | \nu,  K_\theta ) = -\frac{n}{2} \log( (\nu-2) \pi) -\frac{1}{2} \log( |K_\theta|) 
+ \log \bigg( \frac{\Gamma (\frac{\nu+n}{2})}{\Gamma(\frac{\nu}{2})} \bigg) 
 - \frac{(\nu + n)}{2} \log \Big(1+\frac{\beta}{\nu-2} \Big),
\end{align*}

where $\beta = (\boldsymbol{y}-\boldsymbol{\phi})^\top K_\theta^{-1}(\boldsymbol{y}-\boldsymbol{\phi})$
and its derivative with respect to a hyperparameter is
\begin{align*}
\frac{\partial}{\partial \theta} & \log  p(\boldsymbol{y} | \nu, \boldsymbol{\phi}, K_\theta ) = \frac{1}{2} \mathrm{Tr}\bigg(
\Big( \frac{\nu +n }{\nu + \beta -2} \boldsymbol{\alpha} \boldsymbol{\alpha}^\top - K^{-1}_\theta \Big) 
\frac{\partial K_\theta}{\partial \theta} \bigg) , 
\end{align*} 

where $\boldsymbol{\alpha}  = K^{-1}_\theta (\boldsymbol{y}-\boldsymbol{\phi})$. We may also learn $\nu$ using gradient based methods and the following derivative
\begin{align}
\frac{\partial}{\partial \nu}  \log  p(\boldsymbol{y} | \nu, K_\theta ) 
= &-\frac{n}{2(\nu-2)} + \psi \Big( \frac{\nu+n}{2}  \Big) - \psi \Big( \frac{\nu}{2} \Big) \notag \\
& 
- \frac{1}{2} \log \Big(1+\frac{\beta}{\nu-2} \Big) +
 \frac{ (\nu+n)\beta}{2(\nu-2)^2 + 2\beta(\nu-2)}
\end{align} 

where $\psi$ is the digamma function.

\section{More Insight Into the Inverse Wishart Process and Inverse Gamma Priors}
\label{equiv}

As a reminder, we define a Wishart distribution as follows 

{\bf{Definition.}} A random $\Sigma \in \Pi(n)$ is \textit{Wishart} distributed with parameters $\nu >n-1$, $K \in \Pi(n)$ and we write $\Sigma \sim \mathrm{W}_n(\nu, K)$ if its density is given by
\begin{equation}
p(\Sigma) = c_n(\nu, K) |\Sigma|^{(\nu-n-1)/2}
\exp{\Big(-\frac{1}{2}\mathrm{Tr} \big(K^{-1} \Sigma \big)\Big)}, 
\end{equation}
where $c_n(\nu, K) = \Big( |K|^{\nu/2} 2^{\nu n/2} \Gamma_n(\nu/2) \Big)^{-1}$ . \\

\subsection{The Multivariate Gamma Function}

The function in the normalizing constant of the Wishart distribution is called the multivariate gamma function and is defined as follows

{\bf{Definition.}} The \textit{multivariate gamma function}, $\Gamma_n(.)$, is a generalization of the gamma function defined as 
\begin{equation}
\Gamma_n(a) = \int_{S>0} |S|^{a-(n+1)/2} \exp \big( -\mathrm{Tr}(S) \big) dS  
\end{equation}
where $S>0$ means $S$ is positive definite.

In the following lemma we illustrate an explicit relationship between the multivariate gamma function and the gamma function.

\begin{nolem}\textbf{[A]}
\begin{equation}\Gamma_n(a) = \pi^{n(n-1)/4} \prod_{j=1}^n \Gamma \big( a + (1 - j )/2 \big)\end{equation}
\begin{proof}
\begin{align}
\Gamma_n(a) &= \int_{S>0} |S|^{a-(n+1)/2} \exp \big( -\mathrm{Tr}(S) \big) dS  \notag \\
&= \int_{S>0} S_{11}^{a-(n+1)/2} \exp\big( -S_{11} \big) |S_{22.1}|^{a-(n+1)/2} \exp \big( -\mathrm{Tr}(S_{22.1}) \big)   \notag \\
&\hspace{50mm} \times\exp \big( -\mathrm{Tr}(S_{21}S_{11}^{-1}S_{12}) \big) dS_{11} dS_{12} dS_{22.1} \notag \\
& =\int_{S_{11}>0} (\pi S_{11})^{(n-1)/2} S_{11}^{a-(n+1)/2} \exp\big( -S_{11} \big) dS_{11}  \notag \\ 
& \hspace{30mm} \times  \int_{S_{22.1}} |S_{22.1}|^{a-(n+1)/2} \exp \big( -\mathrm{Tr}(S_{22.1}) \big) 
dS_{22.1} \notag \\
&= \pi^{(n-1)/2} \Gamma(a) \Gamma_{n-1}(a - 1/2) \notag
\end{align}
This recursive relationship and the fact that $\Gamma_1(b) = \Gamma(b)$ implies
 \begin{align}
\Gamma_n(a) &= \prod_{j=1}^n \pi^{(j-1)/2} \Gamma(a-(j-1)/2)  \notag \\
&= \pi^{n(n-1)/4} \prod_{j=1}^n \Gamma(a+(1-j)/2)  \notag 
\end{align}
which is as required.
\end{proof}
\end{nolem} 

A simple corollary of this result will be key later.

\begin{nocor}\textbf{[B]}
\begin{equation}
\frac{\Gamma_n(a)}{\Gamma_{n}(a-1/2)} = \frac{\Gamma(a)}{\Gamma(a-n/2)}
\end{equation}
\label{cor}
\end{nocor}

\subsection{Two Different Covariance Priors}

The two generative processes we are interested in are 
\begin{align*}
r^{-1} &\sim \Gamma(\nu/2,1/2)  \hspace{20mm} \Omega \sim \mathrm{W}_n(\nu+n-1,K^{-1})  \\
y_1 & \sim \mathrm{N}_n(0,(\nu-2)rK)   \hspace{10.5mm}        y_2 \sim \mathrm{N}(0,(\nu-2)\Omega^{-1})
\end{align*}
where $n \in \mathbb{N}$, $\nu>2$ and $K$ is a $n \times n$ symmetric, positive definite matrix. 

The marginal distribution for $y_1$ is 
\begin{align}
p(y_1) &= \int p(y_1|r)p(r) dr \notag \\
&= \int (2\pi r(\nu-2))^{-n/2} |K|^{-1/2} \exp \Big(-\frac{y_1^\top K^{-1}y_1}{2(\nu-2)r} \Big) r^{-\nu/2-1} \frac{\exp(-1/(2r))}{2^{\nu/2} \Gamma(\nu/2)} dr \notag \\
& = \frac{(2\pi (\nu-2))^{-n/2} |K|^{-1/2}}{2^{\nu/2} \Gamma(\nu/2)}
\int r^{-(\nu+n)/2-1} \exp \Big( - \Big(1+\frac{y_1^\top K^{-1}y_1}{(\nu-2)}  \Big)/2r \Big) dr \notag \\
&=  \frac{(2\pi (\nu-2))^{-n/2} |K|^{-1/2}}{2^{\nu/2} \Gamma(\nu/2)}
\bigg( \Big( 1+\frac{y_1^\top K^{-1}y_1}{(\nu-2)} \Big)/2 \bigg)^{-(\nu+n)/2} \Gamma \big( (\nu+n)/2 \big) \notag \\
&= (\pi (\nu-2))^{-n/2} |K|^{-1/2}
\Big( 1+\frac{y_1^\top K^{-1}y_1}{(\nu-2)} \Big)^{-(\nu+n)/2} \frac{\Gamma \big( (\nu+n)/2 \big)}{\Gamma(\nu/2)}.
\end{align}

The marginal distribution for $y_1$ is 
\begin{align}
p(y_2) &= \int p(y_2|\Omega)p(\Omega) d\Omega \notag \\
&= \int \big(2\pi(\nu-2)\big)^{-n/2} |\Omega|^{1/2} \exp \Big( - \frac{y_2^\top \Omega y_2}{2(\nu-2)} \Big) \notag \\
&\hspace{20mm} \times c_n(\nu+n-1, K^{-1}) |\Omega|^{(\nu-2)/2}
\exp{\Big(-\frac{1}{2}\mathrm{Tr} \big(K \Omega \big)\Big)} d \Omega
\notag \\
&= \big(2\pi(\nu-2)\big)^{-n/2} c_n(\nu+n-1, K^{-1}) \notag \\
&\hspace{20mm} \times
\int  |\Omega|^{(\nu-1)/2} \exp \bigg( -\frac{1}{2}\mathrm{Tr}\Big( \Big( K+\frac{y_2 y_2^\top }{\nu-2}\Big) \Omega \Big) \bigg)  d \Omega  
\notag \\
&= \big(2\pi(\nu-2)\big)^{-n/2} c_n(\nu+n-1, K^{-1}) \notag \\
&\hspace{20mm} \times
c_n\bigg(\nu+n, \Big( K+\frac{y_2 y_2^\top }{\nu-2}\Big)^{-1}\bigg)^{-1}
\notag \\
&= \big(2\pi(\nu-2)\big)^{-n/2} 
\Big( |K|^{-(\nu+n-1)/2} 2^{(\nu+n-1) n/2} \Gamma_n((\nu+n-1)/2) \Big)^{-1}
\notag \\
&\hspace{10mm} \times  |K|^{-(\nu+n)/2}
\Big(1+\frac{y_2^\top K^{-1} y_2 }{\nu-2}\Big)^{-(\nu+n)/2} 2^{(\nu+n)n/2} \Gamma_n((\nu+n)/2)
\notag \\
&= (\pi (\nu-2))^{-n/2} |K|^{-1/2}
\Big( 1+\frac{y_2^\top K^{-1}y_2}{\nu-2} \Big)^{-(\nu+n)/2} \frac{\Gamma_n \big( (\nu+n)/2 \big)}{\Gamma_n \big( (\nu+n-1)/2 \big)}.
\end{align}

Both marginal distributions are equivalent given the result in Corollary B.

\end{document}